\documentclass[runningheads]{llncs}
\usepackage[T1]{fontenc}
\usepackage{xcolor}
\usepackage{graphicx}
\usepackage{subcaption}
\begin{document}
\title{PXGen: A Post-hoc Explainable Method for Generative Models}
%
%

\author{Yen-Lung Huang\inst{1} \and 
Ming-Hsi Weng\inst{1} \and 
Hao-Tsung Yang\inst{1}} 

\authorrunning{Huang et al.}
%
\institute{Department of Computer Science \& Information Engineering, National Central University, Taiwan  
}
\maketitle              
\begin{abstract}
With the rapid growth of generative AI in numerous applications, explainable AI (XAI) plays a crucial role in ensuring the responsible development and deployment of generative AI technologies. XAI has undergone notable advancements and widespread adoption in recent years, reflecting a concerted push to enhance the transparency, interpretability, and credibility of AI systems. Recent research emphasizes that a proficient XAI method should adhere to a set of criteria, primarily focusing on two key areas. Firstly, it should ensure the quality and fluidity of explanations, encompassing aspects like faithfulness, plausibility, completeness, and tailoring to individual needs. Secondly, the design principle of the XAI system or mechanism should cover the following factors such as reliability, resilience, the verifiability of its outputs, and the transparency of its algorithm. However, research in XAI for generative models remains relatively scarce, with little exploration into how such methods can effectively meet these criteria in that domain.

In this work, we propose PXGen, a post-hoc explainable method for generative models. Given a model that needs to be explained, PXGen prepares two materials for the explanation, the \textit{Anchor set} and \textit{intrinsic \& extrinsic criteria}. Those materials are customizable by users according to their purpose and requirements. Via the calculation of each criterion, each anchor has a set of feature values and PXGen provides example-based explanation methods according to the feature values among all the anchors and illustrated and visualized to the users via tractable algorithms such as $k$-dispersion or $k$-center. Under this framework, PXGen addresses the abovementioned desiderata and provides additional benefits with low execution time, no additional access requirement, etc. Our evaluation shows that PXGen can find representative training samples well compared with the state-of-the-art. 


\keywords{XAI  \and generative AI \and VAE \and post-hoc explanation.}
\end{abstract}
\section{Introduction}

\label{sec: Introduction}
In the field of Artificial Intelligence, generative models such as GANs~\cite{goodfellow2014generative,voynov2020unsupervised}, VAEs~\cite{kingma2013auto,higgins2017beta,lucas2019don}, and diffusion models~\cite{luo2022understanding,ho2020denoising} has been developed well and have a variety of applications including synthetic images \& audio~\cite{burlina2019assessment,donahue2018adversarial}, text generation\cite{gatt2018survey}, reinforcement learning~\cite{tirinzoni2020sequential}, graphic rendering~\cite{tewari2020state} and texture generation~\cite{gao2022get3d}, medical drug synthesis~\cite{walters2020assessing}, $\cdots$ etc. On the other hand, explainable AI (XAI) also rises to dig out the black box of these models. The need for explanation in deep generative models involves many aspects. One intuition is that it allows users or experts to track the generated objects and then control the model behavior in advance. This is crucial especially when the applications are closely related to human safety and moral responsibility.  We need to identify the weaknesses of these applications to improve transparency~\cite{zhou2023explain}. For example, a recent deep learning model restores a blurred image of Obama to a white man only, which implies racial discrimination~\cite{hamilton2020ai}.

\paragraph{Our contribution:}  To address the abovementioned issues, we propose PXGen, a post-hoc explainable method for generative models. Given a model that needs to be explained, PXGen prepares two materials for the explanation, the \textit{Anchor set}, which is a set of samples, and \textit{intrinsic \& extrinsic criteria}, which are two sets of functions to quantify the interaction between the model and a certain anchor. The intrinsic criteria describe the interaction through the latent space of the model, which can be treated as the concept recognition from the model's perspective. The extrinsic criteria, on the other hand, are to describe the difference in outward appearance between the anchors and the outputs of the model. 
Those materials are customizable by users according to their purpose and requirements. Via the calculation of each criterion, each anchor has a set of values which are called feature values. PXGen provides example-based explanation methods according to the feature values among all the anchors such as statistical classification and analysis in advance. Lastly, this analysis is illustrated and visualized to the users via tractable algorithms such as $k$-dispersion or $k$-center. With the path of this framework, PXGen provides an XAI method for general generative models. In two of our demonstration, we presented two model behaviors, model delusion and aligned conception, via our analysis.

To the best of our knowledge, this is the first work that proposes a post-hoc XAI method for black-box encoder-decoder-based generative models. Our approach is highly motivated by recent research which has extensively discussed principles and desiderata of explanations~\cite{schneider2024explainable,lyu2024towards,schneider2019personalized,gunning2019xai}. We selected the most important and well-known desiderata and addressed them in the following.

\begin{itemize}
    \item Faithfulness: The explanation should accurately mirror how the model's process unfolds. The design of intrinsic criteria in PXGen can directly address this property.
    
    \item Robustness: The explanation is verifiable and tractable. All the criteria and algorithms we designed in PXGen are tractable and provide statistical meanings such that one can easily understand and verify the results.
    
    \item Plausibility and Completeness: The explanation should be understandable and contain all relevant factors of the model. Since PXGen is the example-based method this desideratum is highly related to the anchor set and criteria and personalization to the users. 
\end{itemize}

\begin{figure}[t]
     \centering
     \begin{subfigure}[b]{0.5\columnwidth}
         \centering
         \includegraphics[width=\textwidth]{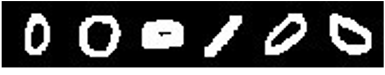}
         \caption{PXGen: Agreeable samples.}
         \label{fig:k-disp on model M}
     \end{subfigure}
     \hfill
     \centering
     \begin{subfigure}[b]{0.5\columnwidth}         
     \centering
         \includegraphics[width=\textwidth]{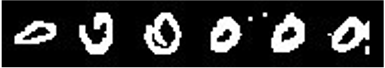}
         \caption{VAE-TracIn: Agreeable samples.}
         \label{fig:k-disp on model M}
     \end{subfigure}
     \hfill
        \caption{The generative model is trained by pictures of the handwriting digit $0$ in MNIST. The two methods (PXGen, VAE-TracIn) are used to find the top-6 agreeable samples, i.e., representative samples of the model.}
       \label{fig:k-disp_vs_tracin}
\end{figure}

Despite the desiderata addressed above, there are a few properties in PXGen we want to emphasize. First, PXGen allows multi-criteria input such that one can have a full explanation covering different aspects. Secondly, PXGen needs no additional access and no interference during the training phase of the model. The calculation time is $O(n^2)$, where $n$ is the size of the anchor set, which is low compared with other non-post-hoc methods such as VAE-TracIn\cite{kong2021understanding}. Lastly, PXGen treats the model as a black-box, this allows PXGen can be applied to a wide range of generative models such as GANs, VAEs, DDPMs, $\cdots$, etc. 

A toy example of PXGen is shown in Figure~\ref{fig:k-disp_vs_tracin}. The generative model is trained with the MNIST dataset~\cite{deng2012mnist}, in which we use pictures of the digit ``0'' only. After that, we use PXGen and VAE-TracIn to find the top six agreeable samples. Compared with VAE-TracIn, PXGen shows a diverse of representative samples based on tractable multi-criteria such that users can have more information and a comprehensive understanding. Besides, our method does not need to access the training phase of the model and the running time is at least one hundred times faster than VAE-TracIn.

In the rest of the paper, section~\ref{sec: Related Work} introduces the related work, where we list various techniques related to XAI. Section~\ref{sec: Framework} presents our PXGen framework, which can discover characteristic Anchors. In section~\ref{sec: DemonstrationVAE}, we show that by using the traditional Variational Encoder and Soft-IntroVAE as the model, we can find our Anchors via our PXGen framework and  evaluate the PXGen framework against other baselines.

\section{Related Work}

\label{sec: Related Work}

With the rise of deep learning models, XAI has grown fast in recent years~\cite{dwivedi2023explainable,hammoudeh2022training,yeh2018representer,guo2020fastif,pruthi2020estimating} and common algorithms such as LIME and SHAP~\cite{ribeiro2016should,lundberg2017unified} have been applied in many learning models. In generative AI, some work is proposed to increase the interpretability of the model. One famous instance is the disentangled method~\cite{tran2017disentangled}, which captures the disentangled information such as pose or angles and separates it into independent factors in the network to learn the hidden representation. Another common method is to calculate the influence function among (training) samples and identify the most responsible samples during the training phase. VAE-TracIn\cite{kong2021understanding}, for instance, measures the significance of individual instances for a specific target by tracking the change in the gradient of the loss during the training process. However, there are only a very limited number of works that address the post-hoc explainable methods in generative AI, which are the ones that do not involve and access the training phase of the model. We emphasize that this is critical since many generative models are from big companies such as OpenAI and Google where access to the training phase is prohibited. Besides, commercial models nowadays are usually trained with billions of data and involve thousands of training epochs. It is unrealistic, or a great effort at least, to trace all the training samples. Another challenge is that, unlike XAI for discriminative models where the data is commonly labeled and has fine criteria with model performance such as accuracy or F1-score. In generative models, the scenario usually involves complex output such as images, audio, tone, style and so on~\cite{yin2022interpreting}. The explanations are also hard to evaluate since the predefined benchmark may not exist or be difficult to create for certain tasks~\cite{schneider2024explainable}.

\section{Framework}
\label{sec: Framework}
\begin{figure}[t]
\centering
\includegraphics[width=0.65\columnwidth]{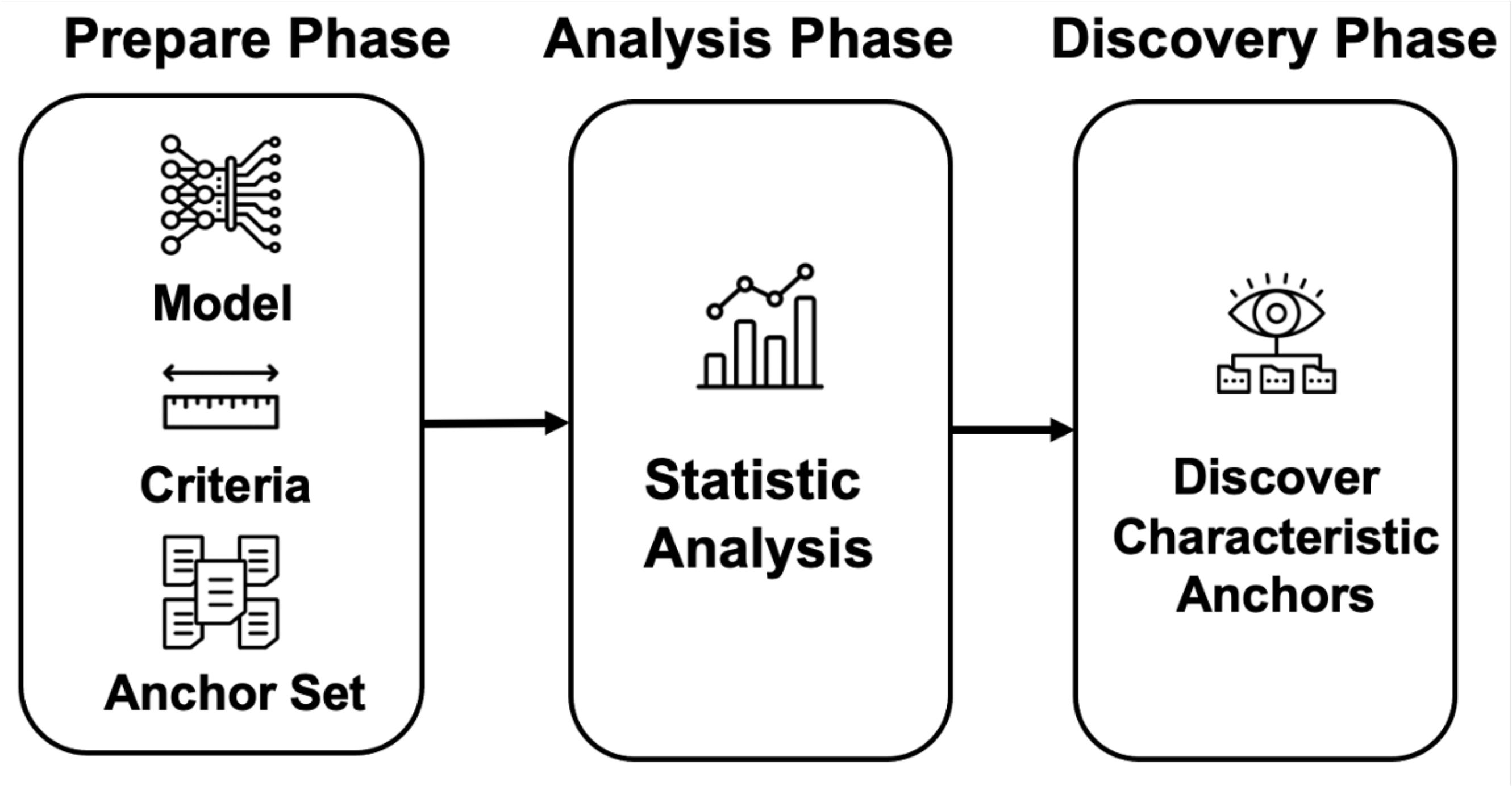}
\caption{The flowchart of PXGen. In the preparation phase, three items, model, criteria, and anchor set are prepared. In the analysis phase, statistical analysis and clas- sification among the anchors are examined to obtain groups with distinct characteristics. In the discovery phase, we identify anchors within specific groups that can provide key information to achieve the explanation.
}
\label{framework}
\end{figure}

In PXGen, the framework is composed of three phrases: Preparation phrase, Analysis phrase, and Discovery phrase. In the Preparation phrase, three objects are needed as the input: the model which is desired to be explained, the anchor set which includes all the instances that are used for explaination, and the (multi-)criteria to describe the various properties of a sample. In  the Analysis phrase, each anchor (i.e., the instance in the anchor set) derives a vector of feature values which are calculated via the criteria. After that, A statistical analysis is provided for post-hoc interpretation of the model. Lastly, we use several algorithms to discover a (sub)set of the anchors and select representative anchors to increase the understandability and visualize the result directly to the users. In the following, we give the details of these three phrases.

To guarantee the faithfulness and robustness of PXGen, all the sub-methods or sub-algorithms in PXGen are tractable and transparent such that one can trace from the result to the origin of the data and the analysis is also repeatable. In the following, we give the details of these three phrases.

\subsection{Preparation Phase}

Initially, three necessary items have to be prepared (model, Anchor set, criteria) for PXGen. 

\paragraph{Model:} The input models PXGen considers are basically the ones that learn the underlying probability distribution of the data they are trained on, enabling them to generate new samples that resemble the original data distribution. Let $\mathcal{X}$ represents the observed data and $\mathcal{Z}$ represent any latent variables. The goal is to learn the conditional distribution $P(\mathcal{X}|\mathcal{Z})$ and the prior distribution $P(\mathcal{Z})$. This allows us can explore both the exterior samples $\mathcal{X}$ and interior prior distribution $\mathcal{Z}$ via different criteria.

\paragraph{Anchor set:} Each anchor is an instance that is formatted consistently with the model's output, such as images or text, but not limited to the training data. As we consider the model to be a black box, we require dataset that can be intuitively understood by people to enter the model for exploration, thus obtaining information about the model's behaviour, which we refer to as the Anchor set. By interacting with the model using these Anchors set to obtain information, different Anchors will yield different information, which can be processed for more concrete results. Through the interaction between Anchors and the model, the model’s behavior is indirectly mapped, Although there is no explicit limitation on the anchors. The selected anchors influence the quality of understandability, completeness, and complexity of the explanation.   

\paragraph{Criteria:} The selected criteria are interpretable functions to process the information resulting from the interaction between Anchors and the model. These criteria should cover different interpretable aspects to acquire varied properties of the plausibility of the anchors. In this work, we divided them into \textit{intrinsic criteria} and \textit{extrinsic criteria}. The intrinsic criteria is to illustrate the ``distance'' between an anchor and the model from the model's perspective. Such as calculating Kullback-Leibler divergence (KLD)~\cite{kullback1951information} of the VAE's\cite{kingma2013auto} latent vector . On the other hand, the extrinsic criteria describes the extrinsic feature between the anchors and the output instances that are generated by the models. Representative examples such as Mean Square Error(MSE)~\cite{bauer1999empirical} and Structural Similarity(SSIM)~\cite{wang2004image}, both are commonly used to measure the similarity between two images.

Based on this observation, we propose a Post-hoc global explainable framework that utilizes real data along with various criteria to explore models. Since we view the model as a black box, we rely on real data as our 'Anchor' points. By classifying using multiple criteria this real data into different groups, we let the real data have various properties,we rely on this approach of indirect mapping to explore and understand why the model generates certain samples. We outline the a framework as follows:

\subsection{Analysis Phase}

In PXGen, anchors intuitively represent the behavior of the model, but not all anchors can adequately represent the model. For example, there should be significant differences in the similarity between the model's output and the original image in the interaction between the model that can only generate dog images and the anchor with horse images. To address this issue, each anchor can be mapped to a multi-dimensional space based on values calculated from \textit{intrinsic } and \textit{extrinsic } criteria, providing specific interpretations in the multi-dimensional space. Another part involves obtaining information from the model and mapping it to the same multi-dimensional space as the anchors, utilizing some statistical analysis methods (such as median, mean, variance, etc.) to analyze the possible distribution of the model from this information. Therefore, users can classify the anchors in the multi-dimensional space to find groups of anchors with unique characteristics, such as those that best represent the behavior of the model, or those that the model erroneously believes are similar to the training data, thereby interpreting the model.


\subsection{Discovery Phase} 

Based on the results of the analysis phase, we can obtain several groups of anchors with different features. For the groups of interest to us (such as the most representative anchor set), we often find that the number is huge, making it difficult for humans to intuitively understand. To overcome this problem, we need some algorithms to help select anchors, resulting in a few key anchors that can represent the particular group. Different algorithms, based on various methods and concepts, would select different Anchors. For example, the k-dispersion\cite{hassin1997approximation,cevallos2015max,cevallos2017local} algorithm picks the most dissimilar individuals in the group, or k-means \cite{vattani2009k} selects the central individuals of the group. User can choose different algorithms according to their needs.



\section{Demonstration : Variational AutoEncode }
\label{sec: DemonstrationVAE}
In this section, we show how to use our framework to identify groups of Anchors with different characteristics, and how to apply algorithms to a specific group to find the most characteristic individuals, achieving an intuitive understanding. The demonstration uses two models, the classical Variational AutoEncoder (VAE) \cite{doersch2016tutorial,kingma2013auto}, and Soft-IntroVAE~\cite{daniel2021soft} as the explained model in PXGen.



\begin{figure}[t]
\centering
\includegraphics[width=0.7\columnwidth]{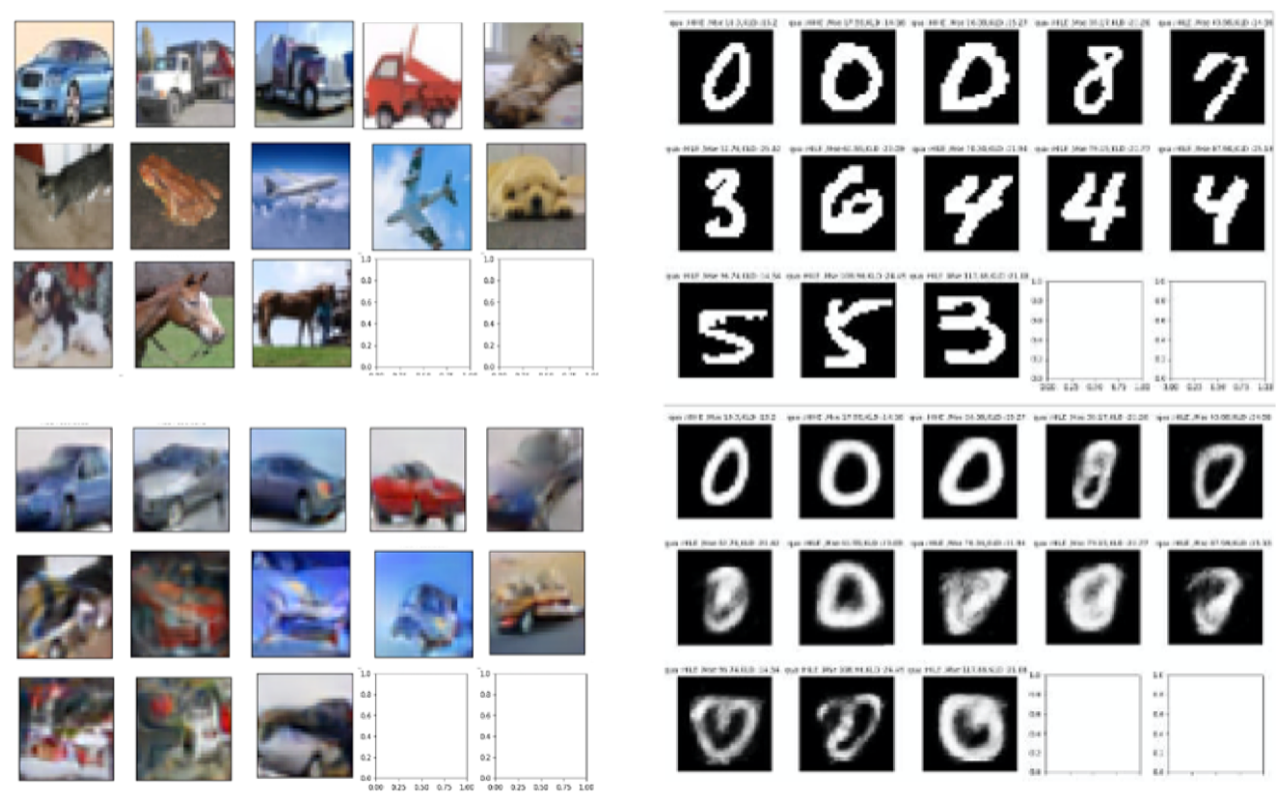}
\caption{In HILE, the phenomenon of ``model delusion'' is displayed, where those anchors are incorrectly decoded. (Top : Original data ; Bottom : Reconstructive data)}
\label{0_10_KLD_quadrantChart}
\end{figure}





\begin{figure}[t]
\centering
\includegraphics[width=0.7\columnwidth]{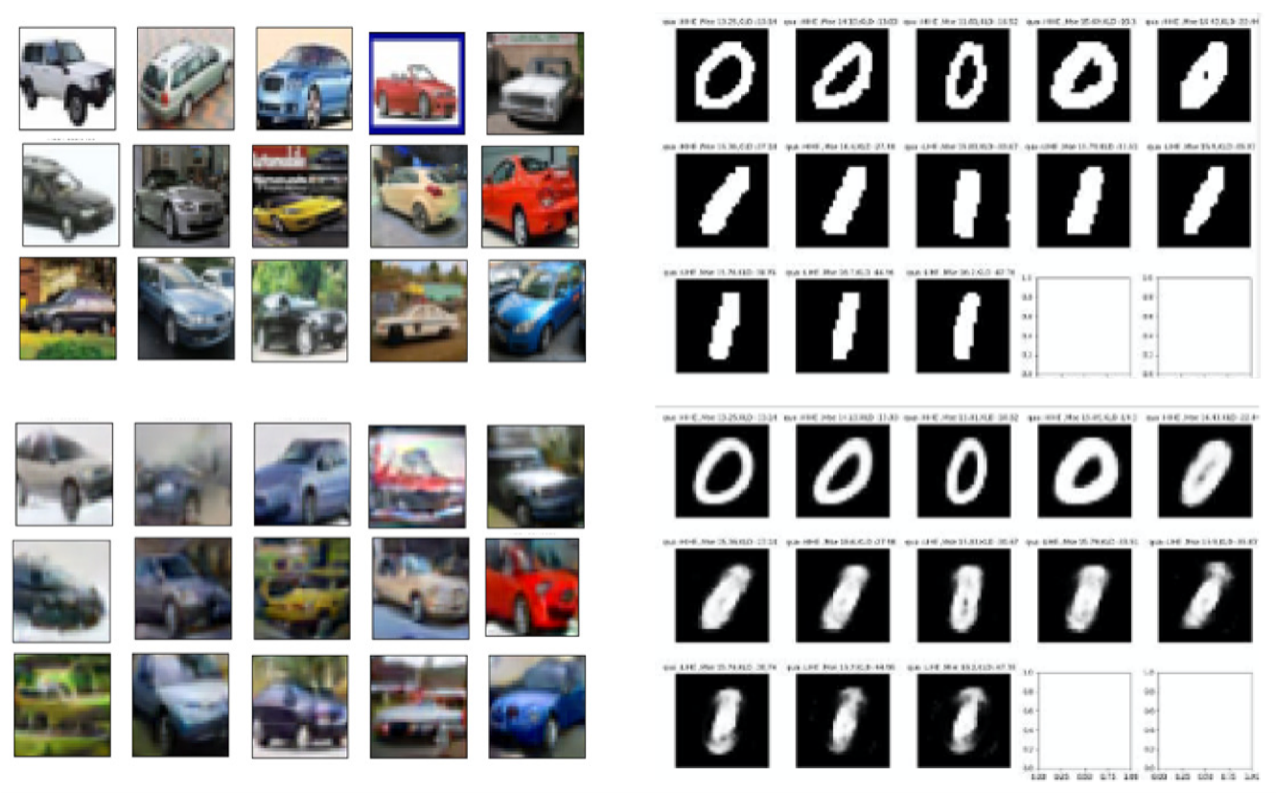}
\caption{In LIHE, a phenomenon of ``aligned conception'' between the handwriting digit “0” and “1” is observed, indicating that there is some agreement of the concept in these two types of samples.
(Top : Original data ; Bottom : Reconstructive data)}
\label{0_15_MSE_quadrantChart}
\end{figure}

\subsection{The Model, Anchor Set, and criteria}
\label{4_Prepare Phase}

The model $M$ is a classical VAE and trained by the hand-writing dataset, MNIST~\cite{deng2012mnist}. To show the explainability of PXGen, we only use  4,000 images with the label ``0'' as the training set. Notice that PXGen has no information of the training process in $M$ nor the training set. The rest of the data in MNIST is used as the anchor set, which is around 54,000 pictures with handwriting digits from ``1'' to ``9'' and approximately 2000 pictures with digit ``0''.


In the multi-criteria part, we choose Kullback-Leibler Divergence (KLD) as the intrinsic criterion and Mean Squared Error (MSE) as the extrinsic criterion. In KLD part, since the generation mechanism of $M$ is based on sampling from a specific distribution. To estimate the predicted distribution of each anchor we reversed the process of $M$, i.e., put the anchor into the encoder to get the predicted distribution from $M$. Then we use KLD to calculate the difference of distribution between the predicted distribution and the actual sampling distribution, $N(0,I)$, of $M$. On the other hand, as we mentioned earlier, MSE is a common criterion  for measure the difference between two simple pictures (in this case, hand writing digits are relatively simple). In calculation, given an anchor $a$, we get the reconstructed sample $a'$ via putting $a$ into the encoder and  decoder sequentially. We then calculate the MSE between $a,a'$. The high-level idea is that, if the MSE is low it means $M$ ``knows'' how to generate pictures similar to $a$. Thus, we can quantify how much $M$ ``knows
'' $a$ via this extrinsic criterion.

Model $M'$ is a Soft-IntroVAE model trained on the CIFAR10 dataset, in which we use 5,000 images with the ``Automobile'' category as the training set. Then, we pick 300 images from each category of the testing dataset as the anchor set. In the multi-criteria part, we choose Kullback-Leibler Divergence (KLD) as the intrinsic criterion. We choose Fréchet Inception Distance (FID)~\cite{heusel2017gans} as the extrinsic criterion, since it may provide a better measurement of the difference between two complex pictures.

\subsection{Analysis Phase: Classifying Anchors and Analysis}
\label{subsec: VAE_4_ClassifyPhase}




To acquire distributions under both extrinsic and intrinsic criteria for the classification of anchors, we first generate a set of images from $M$ . After that, we calculated the maximum values under each criterion as the thresholds. To preserve the stability of the maximum values for each criterion, we used the average of the maximum values obtained from multiple iterations as the thresholds.
There are two advantages of this method. Firstly, it has a statistical context which is not only tractable but also improves the reliability of the threshold. This allows us to associate the threshold with their respective statistical variables more confidently. Secondly, it is independent to the anchor set, meaning that changes to the anchor set would not affect the behavior of the model, which makes PXGen more robust.

With the thresholds, an anchor $a$ is said to have high intrinsic affinity if and only if the KLD value of $a$ is under the average maximum value. Otherwise, it is said to have low intrinsic affinity. Correspondingly, an anchor $a$ is said to have high extrinsic affinity if and only if the MSE value of $a$ is under the average maximum value, and low extrinsic affinity otherwise. By combining these four descriptions, the anchor set can be divided into four groups: HIHE (high intrinsic \& high extrinsic affinity), HILE (high intrinsic \& low extrinsic affinity), LIHE (low intrinsic \& high extrinsic affinity), and LILE (low intrinsic \& low extrinsic affinity).


\paragraph{Analysis:} We have a keen interest in two distinct groups within HILE and LIHE, denoted as ``model delusion'' and ``aligned conception'', respectively. In HILE, anchors exhibit high intrinsic affinity, indicating the model's confidence in generating them with a high probability. However, their low extrinsic affinity suggests that the resulting output image differs significantly from the anchor. We refer to this phenomenon as ``model delusion'', signifying the model's erroneous belief that it understands the ``concept'' of the anchors (i.e., generating a similar latent vector), but misapplies it, resulting in dissimilar outputs after decoding. Concrete examples illustrating this phenomenon are depicted in Figure~\ref{0_10_KLD_quadrantChart}.

In the statistical context, we focus solely on the subset of anchors with a KLD value within the 5\% range among all anchors. We assume that model $M$ correctly identifies the ``concepts'' within this subset of anchors. MSE and FID serve as extrinsic criterion, aiding in quantifying the disparities between the model and external concepts. Therefore, an increase in MSE (or FID) indicates a decrease in extrinsic affinity, reflecting a worsening case of model delusion, as depicted in the top figure. These data points are visualized in the lower portion of Figure~\ref{0_10_KLD_quadrantChart}, where the bottom left figure displays input samples, and the bottom right figure showcases reconstructed samples, sorted from low to high MSE (or FID). It's evident that as extrinsic affinity decreases , the images are erroneously decoded into outputs divergent from their original anchors but aligned with the model's own concept, resembling handwritten digit $0$.

On the contrary, LIHE shows a different story. The model appears to diverge from the ``concept'' of the anchors but effectively utilizes them, a phenomenon we term ``aligned conception''. This suggests that despite differing model concepts, there is alignment to some extent, facilitating transfer learning within this group. Concrete examples are depicted in Figure~\ref{0_15_MSE_quadrantChart}. 

Similarly, the top figure exclusively features anchors with MSE (or FID) values within the 5\% threshold among all anchors, indicating high external affinity with the model. These anchors are sorted based on increasing KLD (decreasing intrinsic affinity), demonstrating alignment with external concepts. The corresponding data points are visually represented in the bottom section of Figure~\ref{0_15_MSE_quadrantChart}. The bottom left figure displays input samples, while the bottom right figure showcases reconstructed samples, ordered from low to high KLD. Remarkably, despite decreasing intrinsic affinity (increasing KLD), the images continue to be accurately decoded into outputs resembling their originals, even without prior exposure to these images.

\begin{figure}[t]
\centering
\includegraphics[width=0.8\columnwidth]{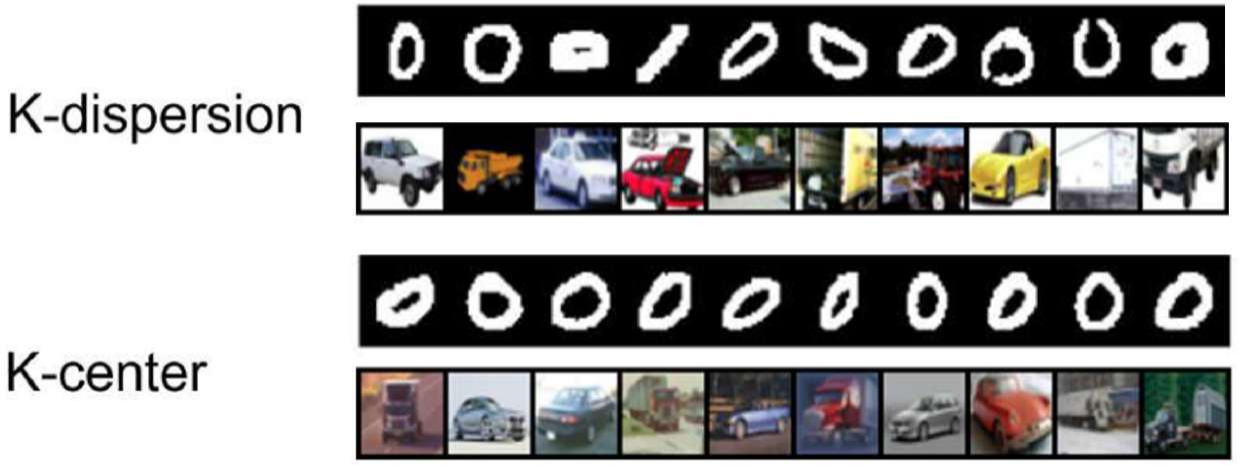}
\caption{Utilizing two algorithms to identify representative anchors within HIHE.}
\label{visualization_all_algorithm}
\end{figure}

\begin{figure}[t]
\centering
\includegraphics[width=0.8\columnwidth]{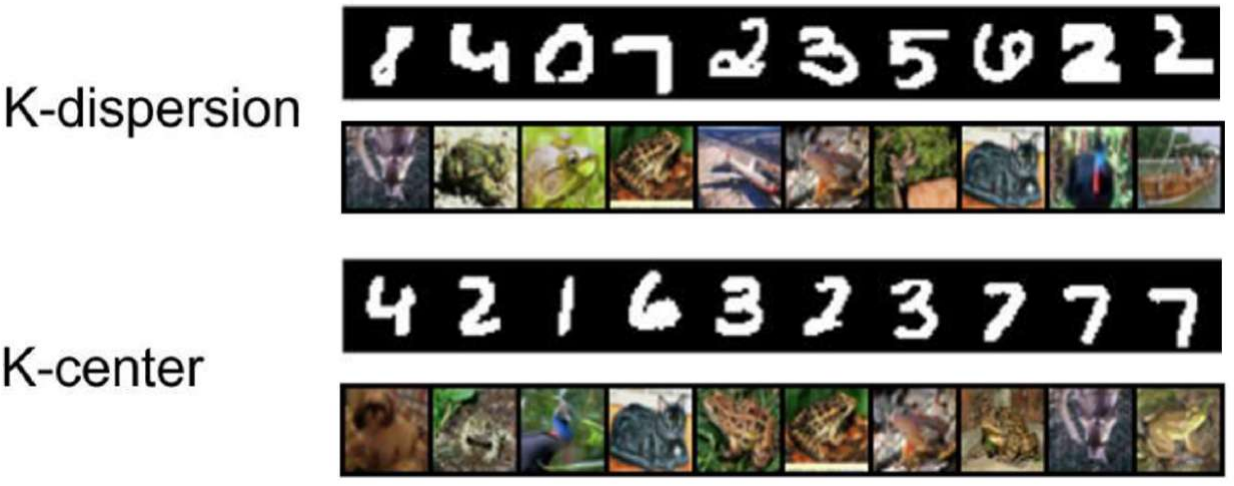}
\caption{Utilizing two algorithms to identify representative anchors within LILE.}
\label{visualization_all_algorithm_quadran1}
\end{figure}

\subsection{Discover Phase: Visualizing the most Characteristic Anchors}
\label{4_DiscoverPhase}



To visualize a group of anchors, PXGen introduces two algorithms for selecting representatives from these group: $k$-dispersion \cite{hassin1997approximation,cevallos2015max,cevallos2017local},  and $k$-center\cite{mentzer2016approximability,chen2021mentzer}. Each of these algorithms considers different concepts. The $k$-dispersion algorithm is designed to identify the $k$ most distinct individuals within a designated group by systematically selecting those $k$ individuals that exhibit the most distinguishing features. And the $k$-center algorithm is an approach used in cluster analysis, which is focused on partitioning a dataset into $k$ distinct clusters. 




In figures \ref{visualization_all_algorithm} and \ref{visualization_all_algorithm_quadran1}, we have demonstrated the results in LILE and HIHE with $k$-dispersion and $k$-center algorithms, with $k=10$. The Anchors in HIHE best represent the possible outputs that the model could generate, whereas the Anchors in LILE are the opposite. Both algorithms offer an intuitive visual understanding of these anchors under different concepts.  


\begin{figure*}[t]
\centering
\includegraphics[width=0.9\textwidth]{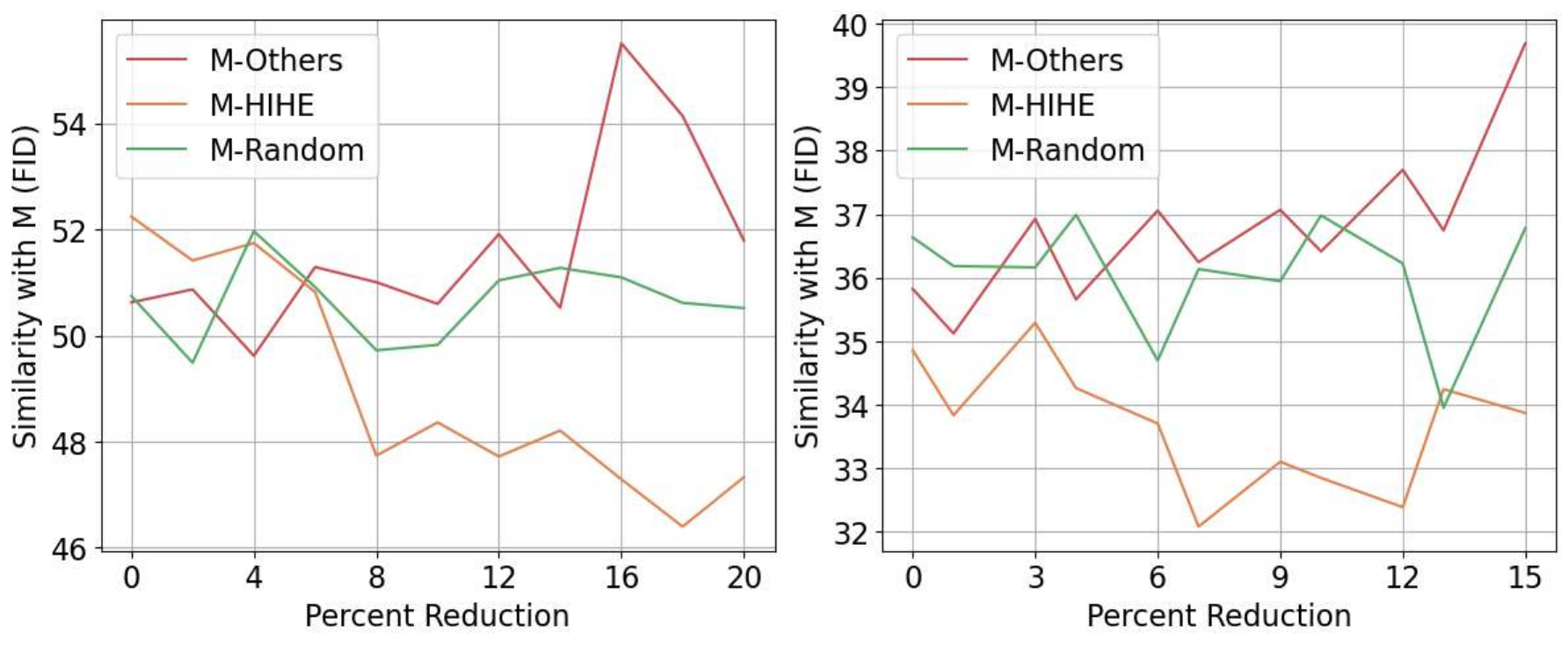} 
\caption{Left: Comparing the similarity results generated by comparing the model trained with class ``0'' to the experimental model. Right: Another experiment trained the model with class ``5''. Both figures show similar trends in different removal processes. }
\label{evaluate_label0_and_5}
\end{figure*}

\begin{figure}[t]
\centering
\includegraphics[width=0.5\columnwidth]{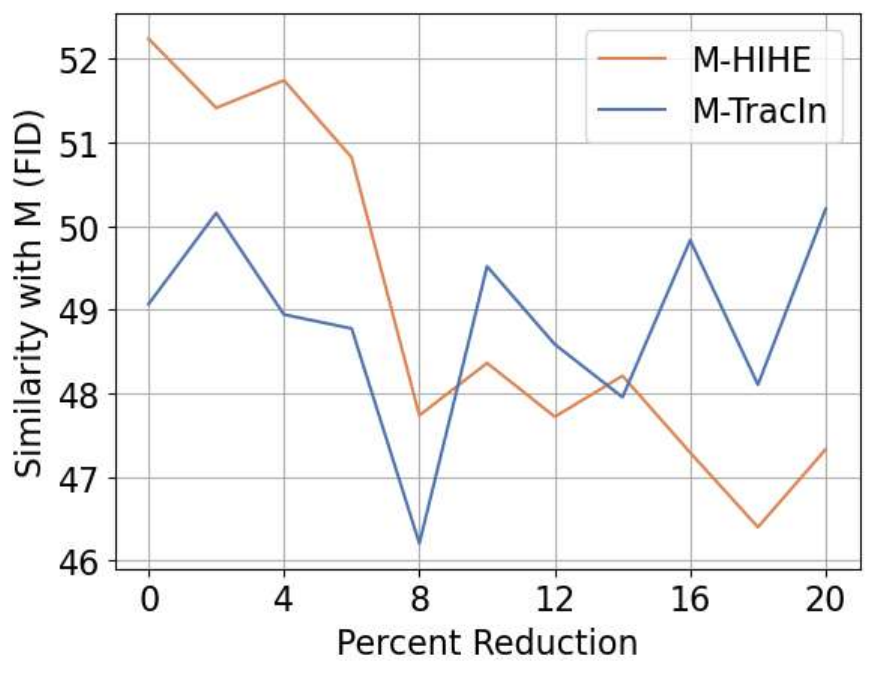}
\caption{We compare our framework with the VAE-TracIn method. by iteratively decrease data to train the models and the generated similarity with the original model. } 
\label{Compare_VAE_TracIn}
\end{figure}

\subsection{Finding Representative Training Samples}
\label{subsec:Finding Representative Training Samples}
Determining the most representative data for the model can be challenginging.  Here, we give a specific purpose for the XAI mission. We want to find a subset of training samples that influences  the trained model the most, or vice versa. That is, the most representative training samples. 

The test model $M$ has the similar setting in Section~\ref{4_Prepare Phase}, which is trained on a training dataset containing 5,923 images with the label ``0''. The criteria for PXGen are  conducted using MSE and KLD and the anchor set is the training dataset only.

Upon utilizing PXGen, the training dataset can be divided into four groups, and each data point is assigned an anchor value coming from the addition of the KLD and MSE values.
According to the discussion in Section~\ref{subsec: VAE_4_ClassifyPhase}, we believe that the training data within the HIHE group are the most representative, while the other groups (LILE, LIHE, HILE) are the opposite. The anchor value indicates the level of representativeness within the group, with lower values being more representative and higher values being less so.

\begin{enumerate}
\item \textit{Validation}
\label{item:Validation}

To verify whether the ``highly representative data'' identified by PXGen, we remove some of the training samples and observe the difference of the model performance after the removal. We compare the image similarity between the original model and the ones with removal. Intuitively, if a model trained with the removal of non-representative samples, the generated results would be closer to the original model, compared with the ones that removes samples randomly. 

We considered three scenarios. Firstly, we trained a model $M$-HIHE using only the data from the HIHE group. Secondly, we trained a model $M$-Others using training data, but removed the training data from the HIHE group in an amount equivalent to that of the other groups, prioritizing the removal of low anchor values. Thirdly, we trained a model $M$-Random by using training data from which an amount equivalent to that of the other groups has been randomly removed, as a baseline. Subsequently, we compared the image similarity between the images generated by these three models and the images generated by the original model . Here, we used the Frechet Inception Distance (FID) \cite{heusel2017gans} to measure image similarity.

In Figure \ref{evaluate_label0_and_5}, we can observe the change in FID values of the three models ($M$-HIHE, $M$-Others, $M$-Random) compared to the original model $M$ as the training data is gradually reduced. It is worth noting that the FID value between $M$-HIHE (which retains only the HIHE data) and model $M$ has decreased, indicating that the data of HIHE  is representative for model $M$.

\item \textit{Comparison with TracIn:}

In this part, we compare PXGen with the VAE-TracIn~\cite{kong2021understanding} method, which evaluates an influence score to determine which training data point is most responsible for increasing the likelihood of a specific data point.  We apply VAE-TracIn to the images generated by model $M$ (3,500 images).  Therefore, for the images generated by Model $M$, each data point in the training dataset of Model $M$ has been assigned 3,500 influence scores. After averaging these scores, we obtain the influence score of this data point for the model, which represents the influence of the data point on the model's generation. In this way, all the training data have an influence score. If the training data has a high influence score, this data point is considered representative.

In addition, we add a model $M$-TracIn is trained by removing the same amount of low-help (harmful) training data. Subsequently, we use $M$-HIHE, $M$-Others, $M$-Random, $M$-TracIn to compare the similarity of generated images (FID) with $M$. In Figure~\ref{Compare_VAE_TracIn}, we individually compared the values of $M$-HIHE and $M$-TracIn, as they are both trained by removing data with low representativeness. Among them, the images generated by $M$-HIHE are more similar to those produced by the original model $M$, indicating that the training data determined by PXGen is more representative.  



\end{enumerate}






\section{Conclusion}
\label{sec: conclusion}
In conclusion, PXGen provides an innovative post-hoc explainable method for generative models by classifying data and recognizing concepts from the model's perspective, aiding in the understanding of complex generative artificial intelligence. Moreover, this method holds great potential for applicability across different scenarios, such as aiding in model training, copyright protection, and more.

\begin{credits}
\subsubsection{\ackname}This work is supported by NSTC: 113-2221-E-008-086.
\end{credits}
%
%
%
\bibliographystyle{splncs04}
\bibliography{mybibliography}

\end{document}